\documentclass[conference]{IEEEtran}

\usepackage{cite}
\usepackage{amsmath,amssymb,amsfonts}
\usepackage{graphicx}
\usepackage{textcomp}
\usepackage{xcolor}
\usepackage{booktabs}
\usepackage{algorithm}
\usepackage{algpseudocode}
\usepackage{amsfonts}
\usepackage{svg}
\usepackage{wrapfig}
\usepackage{array}
\usepackage{float}
\usepackage{placeins}
\usepackage{stfloats}
\usepackage{multirow}
\usepackage{hyperref}

\newcommand\blfootnote[1]{%
  \begingroup
  \renewcommand\thefootnote{}\footnote{#1}%
  \addtocounter{footnote}{-1}%
  \endgroup
}

\def\BibTeX{{\rm B\kern-.05em{\sc i\kern-.025em b}\kern-.08em
    T\kern-.1667em\lower.7ex\hbox{E}\kern-.125emX}}
\begin{document}

\title{MyriadAL: Active Few Shot Learning for Histopathology
}

\author{\IEEEauthorblockN{Nico Schiavone$^\ast$$^\dag$
\hspace{.7cm} Jingyi Wang$^\ast$ \hspace{.7cm} Shuangzhi Li \hspace{.7cm} Roger Zemp \hspace{.7cm} Xingyu Li}
 \IEEEauthorblockA{
 Electrical and Computer Engineering, University of Alberta, Edmonton AB, Canada  \\
 \{nschiavo, jingyi28, shuangzh, rzemp, xingyu\}@ualberta.ca}
}



\maketitle

\blfootnote{This work was supported by the Natural Sciences and Engineering Research Council of Canada (NSERC AOT 567581-21 Zemp, RGPIN-2021-02914 Li), Canadian Institutes of Health Research (CIHR PS 168936 Zemp), UAlberta Huawei-ECE Research Initiative, and Alberta Innovates (LMH 232403439 Zemp, PAB01014 Li). $^\ast$ Equal Contribution. $^\dag$ Corresponding Author.}

\begin{abstract}
Active Learning (AL) and Few Shot Learning (FSL) are two label-efficient methods which have achieved excellent results recently. However, most prior arts in both learning paradigms fail to explore the wealth of the vast unlabelled data. In this study, we address this issue in the scenario where the annotation budget is very limited, yet a large amount of unlabelled data for the target task is available. We frame this work in the context of histopathology image classification, where labelling is prohibitively expensive. To this end, we introduce an active few shot learning framework, Myriad Active Learning (MAL), including a contrastive-learning encoder, pseudo-label generation, and novel query sample selection in the loop. Specifically, we propose to massage unlabelled data in a self-supervised manner, where the obtained data representations and clustering knowledge form the basis to activate the AL loop. With feedback from the oracle in each AL cycle, the pseudo-labels of the unlabelled data are refined by optimizing a shallow task-specific net on top of the encoder. These updated pseudo-labels serve to inform and improve the active learning query selection process. Furthermore, we introduce a novel recipe to combine existing uncertainty measures and utilize the entire uncertainty list to reduce sample redundancy in AL. Extensive experiments on two public histopathology datasets show that MAL has superior test accuracy, macro F1-score, and label efficiency compared to prior works, and can achieve a comparable test accuracy to a fully supervised algorithm while labelling only 5\% of the dataset. Code is available at \url{https://github.com/mesophil/MyriadAL}
\end{abstract}


\section{Introduction}
\label{sec:intro}

Deep learning~\cite{deeplearning} has achieved numerous successes in supervised settings, producing state of the art accuracy and generalization~\cite{dlhisto, dloverview, dlhistobench}. However, for tasks with a scarcity of labelled data, deep learning is not nearly as effective~\cite{deepalplus, whydlnotprod}. 
With the ever increasing amount of unlabelled data, and the growing annotation cost, innovation has shifted towards more label efficient strategies~\cite{breakhis-al, ceal}. In recent years, two label efficient learning paradigms have emerged: Active Learning (AL) and Few Shot Learning (FSL). AL tackles the data scarcity problem by selecting only the most informative data for labelling~\cite{settles2009active}. However, conventional AL models require moderate annotation budgets and often underperform otherwise~\cite{activefewshotlearning}. On the other hand, FSL takes a different approach, utilizing transfer learning or model adaptation/generalization techniques and only a handful of labelled samples from the target dataset~\cite{shakeri2022fhist, fslreview}. 

Despite the promising performance, it should be noted that both AL and FSL utilize only a few pieces of annotated data in model training, leaving the unlabelled samples as untapped potential. We argue that the effective use of the unlabelled data would further improve the performance, especially under the scenario where only a small annotation budget is available. To tackle this challenge, we propose a novel framework for Active Few Shot Learning: Myriad Active Learning (MAL). 
Our first contribution stemming from MAL is a framework that incorporates self-supervised learning, pseudo-labels, and active learning in a positive feedback loop. The pseudo-labels of the unlabelled data are updated every AL cycle and supplement the uncertainty measurement for a more precise and diverse query selection. Our second contribution is designing an algorithm to make the most efficient use of the pseudo-labels in the query selection. The new recipe combines classic uncertainty measures to precisely define sample types based on their comparative uncertainty. We then sample evenly from these types using a self-regulating algorithm, facilitating pseudo-label updates in the next cycle.




We frame the target problem in the context of digital histopathology, where expert pathologists are required to annotate samples in a prohibitively expensive and time-consuming process~\cite{activelearningreviewmedical}. In contrast, the number of unlabelled histopathology images is extremely high, as a single scan can produce hundreds of unique images due to the underlying tissue structures. Under the setting of a very limited annotation budget, we evaluate MAL on two histopathology image sets, 
and show its superiority to classical active learning techniques via comparison and ablation. Notably, MAL can achieve a comparable test accuracy to a fully supervised learning algorithm while labelling only 5\% of a target dataset, demonstrating its potential for effective label efficient learning. 

To the best of our knowledge, no benchmark currently exists for histopathology in the limited budget active learning setting; therefore, MAL also functions as a new benchmark for future works to be compared against. For the purpose of comparing our method against the best available alternatives, we have reworked several well known active learning and few shot learning methods to fit the limited budget setting.

Our contributions are summarized as follows:

\begin{enumerate}
    \item We formulate a new problem: active few-shot learning to address high annotate cost in digital histopathology. The proposed framework utilizes the abundant unlabelled data for more label-efficient model learning.
    \item We develop a novel uncertainty-based active learning algorithm, Myriad Active Learning (MAL). MAL defines a new type of sample diversity, effectively supplementing the existing annotated data for rapid classification accuracy increases.
    \item We show that the proposed method achieves state-of-the-art performance on histopathology datasets in the target setting. As well, with a higher annotation budget, MAL can quickly obtain test accuracies comparable to that of a fully-supervised model.  
\end{enumerate}

\section{Related Work}
\label{sec:relatedwork}


\textbf{Self-Supervised Learning} (SSL) aims to learn generic representations from unlabelled data as a preliminary step to train encoders to efficiently solve subsequent traditional supervised learning tasks~\cite{lecunactivessl}. 
SSL explores data relations by constructing self-supervised pretext tasks based on the unlabeled inputs to produce these representations~\cite{sslcookbook}.
A prominent type of SSL is contrastive learning~\cite{contrastivesslsurvey}, including the recent variant Momentum Contrastive Learning (MoCo) By He et al.~\cite{moco}, which has produced state of the art results on many image-based datasets. SSL algorithms have also produced results competitive with supervised learning on histopathology datasets~\cite{histossl, histosslbasic} when the datasets are large. 
However, most recent efforts to incorporate a limited annotation budget in computational histopathology have been unsuccessful in terms of label efficiency~\cite{histosslbasic, sslskincancer}.


\textbf{Deep Active Learning} (DAL) is another option to tackle the data hungry nature of Deep Learning (DL). The key proposition of DAL is that the majority of the benefits can be obtained from a minority of the samples. In DAL, the training set is unlabelled, optionally with a small number of labelled samples, called the seed. A portion of the unlabelled samples is selected for labelling every cycle, and then included in the training set. 

Sample selection strategies~\cite{deepactivesurvey,deepalplus} are the subject of much innovation, and can be broadly classified into two categories: uncertainty based, and diversity based. Uncertainty based algorithms evaluate samples based on predefined criteria, such as Entropy~\cite{entropysampling}. CEAL ~\cite{ceal}, a recent sampling method, labels the least uncertain samples with their predicted labels, possibly greatly increasing the labelled set size for no additional labelling cost. However, all of these classical tactics underperform expectations on histopathology datasets~\cite{cealhisto, deepalplus}. Diversity sampling methods, such as k-means clustering, or VAAL~\cite{vaal}, posit that the model cannot predict uncertainties accurately enough to be useful. Instead, these methods gather as many types of samples as possible, in order to label, and expose the model to, as many classes as possible. These methods are broadly more effective on histopathology datasets, but are more inconsistent due to their random selection nature within clusters or buckets.



\textbf{Few Shot Learning} (FSL) attempts to tackle the data scarcity problem by learning through only a handful of labelled samples (often $<$1\% of the dataset) in the task domain. This is often accompanied by a well pre-trained model on a large source dataset. FSL has seen great success in histopathology~\cite{shakeri2022fhist}, but is still in a primitive stage with regards to label efficiency research. Active Few Shot Learning provides an potential avenue to solve this problem~\cite{activeoneshotlearningnlp, activefewshotlearning}, but the results in many applications, especially histopathology, have not met expectations. This may be a result of the underperformance of active learning on low annotation budgets in general.





\textbf{Semi-Supervised Learning} (SemiSL) is another method dedicated to maximizing performance on a limited number of labels. Halfway between supervised learning and unsupervised learning, the label information given is often in the form of pre-labelled data, or a randomly selected labelled subset. Active SemiSL has seen success in natural image classification, such as the work by Gao et al.~\cite{gao2020consistencybased}, and in label effective methods utilizing GANs, such as the VAAL method in active learning~\cite{vaal}.

\section{Methods}
\label{sec:methods}

\begin{figure*}[t]
    \centering
    \centerline{\includesvg[inkscapelatex=false,width=1.4\columnwidth]{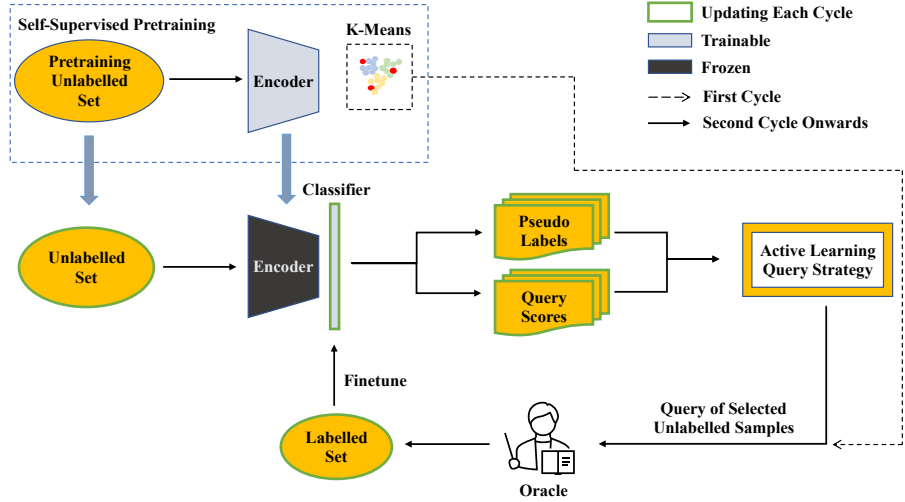}}
    \caption{Diagram of the proposed framework Myriad Active Learning. Pseudo-labels of the unlabelled set are updated and explored for query sample selection.}
    \label{fig:overalldiagram}
\end{figure*}

\subsection{Motivation}
\label{sec:motivation}


In this work, we aim to investigate active few-shot learning solutions under very low annotation budgets. An example of such a scenario is computational histopathology, where annotation budgets are small, and labelling is prohibitively expensive, but unlabelled data is relatively abundant. 

When constructing an effective and efficient solution for this setting, we considered many paradigms and components. We could not utilize many conventional methods, such as knowledge distillation~\cite{iliopoulos2022weighted}, due to the lack of a well-labelled source dataset. Although it is still technically possible to use these methods, they would have to be trained on a more general image dataset, such as ImageNet~\cite{imagenet}. Such a dataset would contain primarily natural images, and this would introduce a large domain gap between the source and target datasets, biasing the model for the downstream tasks.

Therefore, we decided to pursue active-few shot learning, a paradigm which has not yet been explored in the context of histopathology, but has the potential to operate effectively without nearly as many resources.


\subsection{Problem Formulation}
\label{sec:objectives}

Mathematically, the problem can be formulated as follows: we are given a large data set $\mathcal{D} = \{(\textbf{x}_j)\}_{j=1}^{N_{ul}}$ consisting of $N_{ul}$ unlabelled samples $\textbf{x}$ from $K$ categories, and a very small annotation budget. In this case, data sample $\textbf{x} \in R^{H \times W \times C}$ is an image, where $W, H$ are the image width and height, and $C=3$ represents the colour channels. Under the annotation budget, up to $K$ samples can be selected from $D$ for annotation in each active learning cycle. The target is to learn a model $\mathcal{M}$ to predict the categories of queries, i.e., image classification. 

Fig. \ref{fig:overalldiagram} provides a visualization of MAL. Initially, two sets are constructed: one unlabelled set $U_0=D$, and one empty set $\mathcal{T}_0=\{\emptyset\}$ to record the labelled samples from the downstream AL cycles. To explore the rich information in the unlabelled set $D$, we utilize a constrastive learning algorithm to train an encoder, mapping raw histopathology images into numerical features, $f_x=E(x)$. Then, the encoder is frozen and a one-layer classifier is added on top. In the $t^{th}$ active learning cycle, a batch of unlabelled samples $\mathcal{Q}$ are removed from $\mathcal{U}_{t-1}$ for annotation, and the newly labelled samples, $\{(\textbf{x}_i, y_i)\}_{i=1}^{K}$, are added into the labelled set, i.e. $\mathcal{U}_t=\mathcal{U}_{t-1}-\{(\textbf{x}_i)\}_{i=1}^K$ and $\mathcal{T}_t = \mathcal{T}_{t-1}\cup\{(\textbf{x}_i, y_i)\}_{i=1}^K$,  where $y_i \in (0,1)^K$ is the one-hot label vector for $\textbf{x}_i$ over the $K$ classes. The classifier is then optimised on $\mathcal{T}_t$ and generates pseudo-labels $P_t$ for all samples in $\mathcal{U}_{t}$, preparing information for the $t+1^{th}$ cycle. The active learning cycles continue until the budget is exhausted.



\subsection{Self Supervised Learning}
\label{sec:ssl}

In the Myriad Active learning framework, we first train an encoder with self-supervised learning on unlabelled samples in $\mathcal{U}_{0}$. The encoder is then frozen and the extracted numerical features from the unlabelled samples form the basis of the down-stream active learning classification problem. Particularly, the initial clustering effect of SSL encoder helps generate the initial pseudo-labels, which remedies the cold start problem in subsequent active learning. This self-supervision stage is visualized in the top portion of Fig. \ref{fig:overalldiagram}.



For our framework, we utilize the SSL algorithm known as Momentum Contrastive Learning (MoCoV2), by Chen et al.~\cite{mocov2}, which has shown state-of-the-art results on many image-based datasets, including in histopathology~\cite{histosslbasic}. MoCoV2 learns positive/negative (similar/dissimilar) representations from the data, which becomes a list of positive/negative pairs. MoCoV2 formulates this as a dictionary lookup problem, with keys for the representations. Given an unlabelled sample $x \in \mathcal{U}_{0}$, we perform two different data augmentations on $x$ and denote the data augmented versions as $x'$ and $x''$ respectively. Then a query representation $q=f_{x'}=E(x')$ and corresponding key representation $k^+=f_{x''}=E(x'')$ form the positive pair in MoCoV training. Representations from other images constitute a set of negative samples $k^-$. The loss function for encoder optimization is formulated in Eq.~(\ref{eq:mocoloss}).
\begin{multline} \label{eq:mocoloss}
    \mathcal{L}_{q, k^+, \{k^-\}} = \\ - \log{\frac{\exp(q \cdot k^+/\tau)}{\exp(q \cdot k^+ / \tau) + \Sigma_{k^-} \exp(q \cdot k^- / \tau)} }
\end{multline}
where $\tau$ is the temperature hyper-parameter. The large, dynamic dictionary utilized by MoCoV2 is also more efficient than many SSL algorithms~\cite{mocov2}, increasing its usability in real-world settings. 

Note that in this study, we chose not to use more specialized SSL algorithms, such as HistoSSL by Jin et al.~\cite{histossl}, to show that it is not necessary to use a histopathology specific pretraining algorithm for our solution to be effective.

\subsection{Pseudo-Label Generation}
\label{sec:plabels}

Once the SSL model is pretrained on the unlabelled dataset, the learned features are used as an input to a shallow network - a one layer classifier. 
It is important to use a shallow network in this case to reduce the chance of overfitting, as we will only have a few labelled samples per class in the low budget setting.

Initially, there is no training data for the target task, as we do not use an initial seed; therefore, the shallow network cannot provide any meaningful information to the active learning algorithm. Instead, we use K-Means clustering on the numerical features from the frozen encoder to form $K$ clusters in the first cycle. The first query is composed of one sample selected from each of the clusters. From the second cycle onwards, the framework proceeds in a closed-loop fashion, as depicted in Fig. \ref{fig:overalldiagram}. The classifier is updated using the labelled samples in $\mathcal{T}_{t}$, and generates the set of pseudo-labels $P_t$ to use in the next cycle. The pseudo-labels $\hat{y}$ are generated based on the predicted class probabilities for each sample $\textbf{x}_j \in \mathcal{U}_t$; defining $\hat{y}_j$ as the most likely class label of $\textbf{x}_j$, the set of pseudo-labels $P_t = \{\hat{y}_j\}_{j=1}^{N_{ul}}$ is formed. Excellent pseudo-labels are crucial to the function of our framework, as the pseudo-labels will inform the diversity of the active learning query and be the best mode of prevention against redundant samples.

\subsection{Myriad Active Learning}

Myriad Active Learning (MAL) provides a solution to the goal of collecting a diverse query, minimizing redundancy, and maximizing the accuracy in each cycle. To accomplish this, we utilize conventional active learning techniques in a novel way, allowing for more precise and deliberate sample selection. MAL uses a modified version of conventional uncertainty-based strategies, the pseudo-labels obtained using the methods in Sec. \ref{sec:plabels}, and a novel sample selection strategy. First, we combine margin sampling and entropy sampling into Margin-Entropy (M-E) Sampling to allow for a more precise mapping of the uncertainties (using $\mathcal{M}$ to denote the model):

\begin{equation}
    \sigma_{\textbf{m-e}}(\textbf{x}, \mathcal{M}) = \frac{1}{\sigma_{\textbf{margin}}(\textbf{x}, \mathcal{M})} + \sigma_{\textbf{entropy}}(\textbf{x}, \mathcal{M}),
\end{equation}

\begin{equation}
    \sigma_{\textbf{margin}}(\textbf{x}, \mathcal{M}) = p_\mathcal{M}(\hat{y}_{1} | \textbf{x}) - p_\mathcal{M}(\hat{y}_{2} | \textbf{x}),  
\end{equation}

\begin{equation}
    \sigma_{\textbf{entropy}}(\textbf{x}, \mathcal{M}) = - \Sigma_{i=1}^k p_\mathcal{M}(y_i | \textbf{x}) \log{p_\mathcal{M}(y_i | \textbf{x})},
\end{equation}
where $p_\mathcal{M}(\hat{y} | \textbf{x})$ represents the class probability of sample $\textbf{x}$ under model $\mathcal{M}$. The aforementioned classical uncertainty sampling strategies are defined as follows: \textbf{Margin Sampling}~\cite{marginsampling} gives the uncertainty based on the difference between the probabilities of a sample's most and second most likely labels ($\hat{y}_{1}$ and $\hat{y}_{2}$, respectively). In this case, a lower value means a larger uncertainty. \textbf{Entropy Sampling}~\cite{entropysampling} selects data based on the maximal entropy. When combined, the $1/\sigma_{\textbf{margin}}$ is used rather than $\sigma_{\textbf{margin}}$ to ensure larger values map to larger uncertainties. 

Margin sampling favours samples closer to the decision boundaries~\cite{marginsampling}, regardless of how many classes intersect at those boundaries. Entropy sampling favours the most uncertain samples close to the decision boundary of \textit{as many} classes as possible, and therefore often picks out noisy or difficult samples. Classically, the samples will be selected from the most uncertain to the last uncertain until the per cycle quota $K$ is filled.

\begin{figure*}[t]
\centerline{\includesvg[inkscapelatex=false,width=1.4\columnwidth]{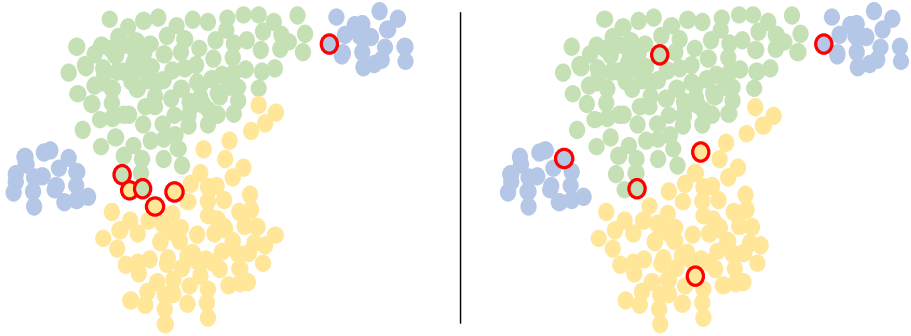}}
\caption{Abstracted t-SNE plot of example data (3-class). Left: 6 samples selected by classical active learning methods with entropy sampling, notably selecting many samples from the same area which are highly likely to be redundant. Right: 6 samples selected using MAL, finetuning several of the borders simultaneously, while providing anchor samples for two of the classes.}
\label{fig:mal}
\end{figure*}

The reason for this is that these sampling methods only provide concrete information on where the \textit{highest} uncertainty samples will be found. As a result, similar samples are often labelled, potentially wasting label information. A visualization of these conventional methods can be found on the left side of Fig. \ref{fig:mal}.

Comparatively, M-E has a much more predictable structure: a low M-E uncertainty sample is likely to be found near the centre of a cluster, due to a low margin score \textit{and} a low entropy. Conversely, a high M-E uncertainty sample will almost certainly be found near a decision boundary. In this way, the entire list of uncertainties can be deliberately utilized. The overconfident predictions at the low end of the list serve to establish anchors, while the top end of the list samples the usual suspects from conventional methods. We utilize these characteristics by splitting up the sorted uncertainty list into $K$ sub-arrays, so that no more than one sample will be selected from the same area. A visual representation of this is given in Fig. \ref{fig:mal}, intuitively showing the difference between MAL and the current techniques.

\begin{algorithm}[t] 
\caption{Myriad Active Learning}\label{alg:myriad}
\begin{algorithmic}
\Require $N_{ul} \geq K$, $K \geq 1$, $t > 1$
\State $\alpha \gets [\quad]$, $\beta \gets [\quad]$, $\mathcal{Q} \gets [\quad]$, $S \gets [\quad]$, $n \gets 0$\\
\textbf{for all} {$\textbf{x}_i \in \mathcal{U}_t$} \textbf{do:} \; \textbf{append:} $\sigma_{\textbf{m-e}}(\textbf{x}_i, \mathcal{M})$ \textbf{onto} $\alpha$ 
\State $\beta$ $\gets$ \textbf{ArgSort:} $\alpha$ in \textbf{descending} order
\State \textbf{Split:} $\beta$ into $K$ approximately equally sized \textbf{subarrays} $\beta_0, ..., \beta_{K-1}$
\While{\textbf{len}($\mathcal{Q}$) $<$ $K$}
\For {$i$ \textbf{in} $\beta_n$}
    \If {$P_t(i)$ \textbf{not in} $S$}
        \State \textbf{append:} $\mathcal{U}_t (i)$ \textbf{onto} $\mathcal{Q}$
        \State \textbf{append:} $P_t (i)$ \textbf{onto} $S$
        \State \textbf{break}
    \EndIf
\EndFor
\State \textbf{if} {$n \geq K-1$} \textbf{then} $n \gets 0$ \textbf{else} $n \gets n + 1$    
\EndWhile

\end{algorithmic}
\end{algorithm}

Combining M-E sampling with the diversity presented by the pseudo-labels, we create the novel selection algorithm Myriad Active Learning (MAL). The pseudocode for MAL can be found in Alg. \ref{alg:myriad}.
A summary of the algorithm follows: MAL takes in $\mathcal{U}_t$, $P_t$, and $K$ as inputs, calculates the uncertainty $\sigma$ for every sample in $\mathcal{U}_t$, and stores them in order in an array $\alpha$. This array is argument sorted into $\beta$, such that $\beta[i]$ is the index of the $i^{th}$ most uncertain sample in $\mathcal{U}_t$. $\beta$ is then split into $K$ sub-arrays $\beta_0, ..., \beta_{K-1}$, and one-by-one the most uncertain sample from each sub-array is added to $\mathcal{Q}$. After each sample is selected, its corresponding pseudo-label from $P_t$ is appended to $S$, and subsequent samples added to $\mathcal{Q}$ must have a pseudo-label which is not in $S$. This is to ensure that a sample with a different pseudo-label is selected from each sub-array. This portion loops until $\mathcal{Q}$ contains $K$ samples, at which point they are sent to the oracle for labelling.

\section{Experiments}
\label{sec:experiments}


\subsection{Datasets}
\label{sec:datasets}

The proposed framework, MAL, is evaluated on two public histopathology datasets. 

\textbf{NCT-CRC-HE-100K} (NCT) ~\cite{nctdataset} provides 100,000 non-overlapping histopathology image patches cropped from 86 H\&E stained tissue slides by the National Center for Tumor Diseases and the University Medical Center Mannheim pathology archive. All images in NCT are color-normalized with 224x224 pixels at 0.5 microns per pixel. The 9 tissue categories included in this dataset are adipose, background, debris, lymphocytes, mucus, smooth muscle, normal colon mucosa, cancer-associated stroma, and colorectal adenocarcinoma epithelium.

\textbf{Breast Cancer Histopathological Database} (BreaKHis) ~\cite{breakhisdataset} is composed of 9,109 H\&E stained breast tumor images from 82 patients, with 2,480 benign and 5,429 malignant samples. Aside from the binary labels of benign and malignant, images in BreakHis are further categorized into 8 categories, which are adenosis, fibroadenoma, phyllodes tumor, and tubular adenona, carcinoma, lobular carcinoma, mucinous carcinoma, and papillary carcinoma. All images in BreaKHis is with 700X460 pixels in the PNG format of 3-Channel RGB, 8-bit depth in each channel. Different from the NCT dataset, BreakHis is a very difficult dataset since it includes images with different magnifying factors (40X, 100X, 200X, and 400X) which makes computational image diagnosis more challenging.


\begin{table*}[t]
\renewcommand{\arraystretch}{1.25}
\centering
\begin{tabular}{c|ccc|ccccc}
\toprule
Dataset & PrT LR & PrT Batch Size & PrT Epochs & LR & Batch Size & Epochs   \\ \midrule
NCT & 0.015 & 128 & 200 & 0.0004 & 128 & 200  \\
BreaKHis & 0.0005 & 32 & 20 & 0.0012 & 128 & 100  \\ \bottomrule
\end{tabular}
\vspace{5pt}
\caption{Hyper-parameters (i.e. learning rate, batch size, and training epoch number) for SSL-pretraining (denoted PrT) encoder and the active learning loop. A weight decay of 0.0005 is used in all cases.}
\end{table*}
\label{table:hyperparams}
\subsection{Experimental Protocol}
\label{sec:protocol}

In our study, NCT and BreaKHis constitute 9-category classification and 8-class diagnosis tasks, respectively. For each dataset, images are divided into unlabelled training set and test set with a ratio of 80:20. The samples were divided randomly, as we operate under the assumption that we do not know the underlying distribution of the datasets (i.e. we have no label information at this stage).

This study focuses on the scenario of active few-shot learning where a very low annotation budget is available. That is, we allow only $n$-shot samples for data annotation in experimentation, where $n=1, 5,  10$. 

The self-supervised learning model is composed of a ResNet-50 backbone and a 2-layer MLP head (2048-dimensional hidden layer with a ReLU activation). We follow the study in Chen et al.~\cite{mocov2} for data augmentation. One exception is that for the BreaKHis dataset, a 460x460-pixel crop was used for the randomly resized images. The one-layer classification network is initialized using Xavier uniform distribution and is optimized with ADAM~\cite{adamoptimizer}. The hyper-parameters including learning rate, batch size, and training epoch number, are specified in Table \ref{table:hyperparams}. 

\subsection{Comparison Baseline}
To the best of our knowledge, no prior works address the active few-shot learning scenario in histopathology, so we will make our comparisons over several different paradigms. 
Specifically, we divide our comparison experiments into two main parts. First, we compare MAL to a recent few-shot learning benchmark, FHIST~\cite{shakeri2022fhist}. Here, we use classification accuracy and macro F1 scores as the performance metrics. Then, we compare our methods to popular active learning methods, summarized by Zhan et al.~\cite{deepalplus}. 

Note that both the few-shot learning results and active learning results reported in this paper were reproduced using the official code published by their respective authors. The active learning methods required an initial seed, as they do not utilize pretraining, so they are given $K$ randomly chosen labelled samples initially. All results are reported with the mean and sample standard deviation from 3 seeds.

\begin{table*}[t]
\renewcommand{\arraystretch}{1.25}
\centering
\begin{tabular}{c|c|ccc|ccc}
\toprule
& & & Accuracy & & & F1 Score & \\ 
\multirow{1}{4em}{Datasets} & \multicolumn{1}{c|}{Method $\downarrow$} & \multicolumn{1}{c}{1-shot} & \multicolumn{1}{c}{5-shot} & \multicolumn{1}{c|}{10-shot} & \multicolumn{1}{c}{1-shot} & \multicolumn{1}{c}{5-shot} & \multicolumn{1}{c}{10-shot} \\ \midrule
\multirow{2}{4em}{NCT} & \multicolumn{1}{c|}{MAL} & 48.7\scriptsize$\pm6.7$ & \textbf{77.9}\scriptsize$\pm2.0$ & \textbf{87.1}\scriptsize$\pm0.5$ & \textbf{51.2}\scriptsize$\pm9.0$ & \textbf{79.5}\scriptsize$\pm2.3$ & \textbf{88.0}\scriptsize$\pm0.4$ \\
& \multicolumn{1}{c|}{FHIST ~\cite{shakeri2022fhist}} & \textbf{56.2}\scriptsize$\pm10.8$ & 75.4\scriptsize$\pm8.1$ & 80.9\scriptsize$\pm7.2$ & 30.3\scriptsize$\pm6.2$ & 41.8\scriptsize$\pm4.5$ & 44.9\scriptsize$\pm4.0$  \\ \midrule

\multirow{2}{4em}{BreaKHis} & \multicolumn{1}{c|}{MAL } & \textbf{33.9}\scriptsize$\pm6.6$ & 51.6\scriptsize$\pm4.0$ & \textbf{65.1}\scriptsize$\pm1.7$ & 16.2\scriptsize$\pm5.5$ & \textbf{29.7}\scriptsize$\pm3.1$ & \textbf{37.2}\scriptsize$\pm0.9$ \\
& \multicolumn{1}{c|}{FHIST ~\cite{shakeri2022fhist}} & 33.8\scriptsize$\pm7.36$ & \textbf{53.1}\scriptsize$\pm8.54$ & 62.0\scriptsize$\pm7.44$ & \textbf{18.0}\scriptsize$\pm4.1$ & 29.1\scriptsize$\pm4.9$ & 34.2\scriptsize$\pm4.2$  \\ \bottomrule

\end{tabular}
\vspace{5pt}
\caption{Test accuracies (\%) and Macro F1 scores (\%) on the NCT and BreaKHis datasets in the few shot learning setting. FHIST by Shakeri et al.~\cite{shakeri2022fhist} is a recent study presenting a few-shot learning benchmark on histopathology images.}
\label{table:fsl}
\end{table*}

\begin{table*}[t]
\renewcommand{\arraystretch}{1.25}
\centering
\begin{tabular}{c|cccccccc}
\toprule
\multirow{1}{4em}{Dataset}  & \multicolumn{1}{l|}{Method \cite{deepalplus} $\rightarrow$} & \multicolumn{1}{c}{MAL} & \multicolumn{1}{c}{Rand.} & \multicolumn{1}{c}{Margin} & \multicolumn{1}{c}{Entropy} & \multicolumn{1}{c}{VarRatio} & \multicolumn{1}{c}{CEAL} & \multicolumn{1}{c}{KMeans} \\ \midrule
\multirow{3}{4em}{NCT} &\multicolumn{1}{c|}{1-shot} & \textbf{48.7}\scriptsize$\pm6.7$  & 29.0\scriptsize$\pm3.3$ & 32.5\scriptsize$\pm1.5$ & 24.5\scriptsize$\pm0.9$ & 27.0\scriptsize$\pm5.0$ & 30.2\scriptsize$\pm2.2$ & 28.4\scriptsize$\pm7.7$ \\
& \multicolumn{1}{c|}{5-shot} & \textbf{77.9}\scriptsize$\pm2.0$  & 35.7\scriptsize$\pm4.1$ & 39.9\scriptsize$\pm4.5$ & 33.7\scriptsize$\pm2.4$ & 32.0\scriptsize$\pm7.7$ & 31.7\scriptsize$\pm10.5$ & 33.7\scriptsize$\pm3.9$ \\
& \multicolumn{1}{c|}{10-shot} & \textbf{87.1}\scriptsize$\pm0.5$  & 47.7\scriptsize$\pm3.8$ & 42.5\scriptsize$\pm2.3$ & 38.0\scriptsize$\pm10.0$ & 42.7\scriptsize$\pm10.0$ & 37.9\scriptsize$\pm5.0$ & 33.7\scriptsize$\pm3.0$ \\ \midrule
\multirow{3}{4em}{BreaKHis} &\multicolumn{1}{c|}{1-shot} & 33.9\scriptsize$\pm6.6$ & 25.6\scriptsize$\pm13.4$ &  38.3\scriptsize$\pm2.8$ & 31.9\scriptsize$\pm7.3$ & 44.3\scriptsize$\pm0.5$ & 39.5\scriptsize$\pm2.1$ & \textbf{43.7}\scriptsize$\pm0.6$ \\
& \multicolumn{1}{c|}{5-shot} & \textbf{51.6}\scriptsize$\pm4.0$  & 44.5\scriptsize$\pm0.9$ & 44.6\scriptsize$\pm0.8$ & 44.6\scriptsize$\pm6.3$ & 45.7\scriptsize$\pm2.2$ & 43.1\scriptsize$\pm2.9$ & 42.6\scriptsize$\pm1.4$ \\
& \multicolumn{1}{c|}{10-shot} & \textbf{65.1}\scriptsize$\pm1.7$  & 43.4\scriptsize$\pm0.9$ & 49.1\scriptsize$\pm0.3$ & 46.1\scriptsize$\pm2.2$ & 48.9\scriptsize$\pm0.8$ & 44.2\scriptsize$\pm2.1$ & 46.2\scriptsize$\pm2.1$ \\ \bottomrule
\end{tabular}
\vspace{5pt}
\caption{Test accuracies (\%) for the NCT and 8-class BreaKHis datasets in the active learning setting. MAL is compared to conventional deep learning methods evaluated in Zhan et al.~\cite{deepalplus}.}
\label{table:al}
\end{table*}

\begin{table*}[t]
\renewcommand{\arraystretch}{1.25}
\centering
\begin{tabular}{c|cccccccc}
\toprule
\multirow{1}{4em}{Dataset}  & \multicolumn{1}{l|}{Method \cite{deepalplus} $\rightarrow$} & \multicolumn{1}{c}{MAL} & \multicolumn{1}{c}{Rand.} & \multicolumn{1}{c}{Margin} & \multicolumn{1}{c}{Entropy} & \multicolumn{1}{c}{VarRatio} & \multicolumn{1}{c}{CEAL} & \multicolumn{1}{c}{KMeans} \\ \midrule
\multirow{3}{4em}{NCT} &\multicolumn{1}{c|}{1-shot} & \textbf{51.2}\scriptsize$\pm9.0$ & 22.0\scriptsize$\pm7.1$ & 24.1\scriptsize$\pm2.5$ & 17.4\scriptsize$\pm2.4$ & 21.3\scriptsize$\pm7.1$ & 23.0\scriptsize$\pm5.0$ & 20.5\scriptsize$\pm8.9$ \\
& \multicolumn{1}{c|}{5-shot} & \textbf{79.5}\scriptsize$\pm2.3$ & 31.3\scriptsize$\pm2.1$ & 36.2\scriptsize$\pm4.7$ & 27.5\scriptsize$\pm3.2$ & 24.6\scriptsize$\pm6.9$ & 25.8\scriptsize$\pm9.8$ & 27.1\scriptsize$\pm3.3$ \\
& \multicolumn{1}{c|}{10-shot} & \textbf{88.0}\scriptsize$\pm0.4$ & 44.5\scriptsize$\pm4.8$ & 40.6\scriptsize$\pm1.8$ & 34.6\scriptsize$\pm12.0$ & 34.6\scriptsize$\pm12.0$ & 34.8\scriptsize$\pm6.1$ & 27.0\scriptsize$\pm3.6$ \\ \midrule
\multirow{3}{4em}{BreaKHis} &\multicolumn{1}{c|}{1-shot} & \textbf{16.2}\scriptsize$\pm5.5$ & 6.0\scriptsize$\pm1.9$ & 11.0\scriptsize$\pm1.2$ & 10.1\scriptsize$\pm2.3$ & 9.3\scriptsize$\pm1.2$ & 9.5\scriptsize$\pm1.0$ & 11.9\scriptsize$\pm1.6$ \\
& \multicolumn{1}{c|}{5-shot} & \textbf{29.7}\scriptsize$\pm3.1$ & 9.1\scriptsize$\pm0.9$ & 10.4\scriptsize$\pm1.3$ & 15.6\scriptsize$\pm1.4$ & 13.0\scriptsize$\pm1.8$ & 13.9\scriptsize$\pm0.6$ & 9.9\scriptsize$\pm1.3$ \\
& \multicolumn{1}{c|}{10-shot} & \textbf{37.2}\scriptsize$\pm0.9$ & 8.9\scriptsize$\pm0.9$ & 15.7\scriptsize$\pm0.7$ & 15.3\scriptsize$\pm0.7$ & 14.3\scriptsize$\pm0.9$ & 14.9\scriptsize$\pm2.2$ & 12.5\scriptsize$\pm1.4$ \\ \bottomrule
\end{tabular}
\vspace{5pt}
\caption{Macro F1 scores (\%) for the NCT and 8-class BreaKHis datasets in the active learning setting. MAL is compared to conventional deep learning methods evaluated in Zhan et al.~\cite{deepalplus}.}
\label{table:al2}
\end{table*}

\subsection{Main Results and Discussion}
\label{sec:mainresults}

\textbf{Few-Shot Learning Comparisons}: FHIST~\cite{shakeri2022fhist} is a recent few shot learning benchmark particularly designed for histopathology images. FHIST uses a large neural network pretrained on a well-annotated pathology image set, which is then transferred to and finetuned with the few-shot samples from the target dataset (NCT and BreaKHis in this case). Table \ref{table:fsl} shows the $n$-shot learning results on NCT/BreaKHis with FHIST and MAL. 

It is reasonable that the 1-shot accuracies are lower, as the classifier in MAL trains only on $K$ samples, or a possible one sample per class from K-means, while FHIST takes advantage of the knowledge transferred from the extra annotated data. However, when more data is selected and annotated, for example, in 5-shots and 10-shots, MAL substantially improves the test accuracy and macro F1 score of the model, and significantly outperforms FHIST. Notably, in all 10-shot cases, MAL outperforms FHIST. 

Higher budget settings are also explored in Table~\ref{table:ssl}. A CNN trained from scratch achieves 96.16\% accuracy on NCT~\cite{ghoshnctcnn}, which can be boosted to 99.76\% using transfer learning~\cite{BENHAMIDA2021104730}. MAL achieves comparable results at only 5\% labels, with a test accuracy of 95.9\%.

\textbf{Active Learning Comparisons}: Next, we compare MAL against other classical deep active learning methods, using the methodology described by Zhan et al.~\cite{deepalplus} on the 8-class BreaKHis dataset and the NCT dataset. As shown in Tables \ref{table:al} and \ref{table:al2}, MAL improves upon popular deep active learning methods in the few shot setting. Specifically, macro F1 score is improved by 4.3\%, 14.1\%, and 21.5\% at 1, 5, and 10-shots, respectively for the BreaKHis dataset. For NCT, a similar trend is observed with a 27.1\%, 43.3\%, and 43.5\% increase in macro F1 score at 1, 5, and 10-shots, respectively. Similar trends are observed for the test accuracies.

\textbf{Higher Budget Settings}: In this setting, we relax the few-shot condition and investigate the performance of MAL with a higher annotation budget and report its performance in Table \ref{table:ssl}. For a reference, a fully supervised CNN-based classifier trained on the entire NCT dataset achieves 96.16\% accuracy \cite{EfficientNet, ghoshnctcnn}. That is, MAL is able to achieve similar performance with only 5\% annotation by always selecting the most informative samples to supplement model learning.

\textbf{Discussion and limitations}: Our experiments show that MAL outperforms prior FSL and AL methods on histopathology images ~\cite{shakeri2022fhist, deepalplus} in terms of accuracy, macro F1 score, and label efficiency.

There is a notable large gap in macro F1 score between MAL and the other FSL and AL methods on the NCT dataset. This is due to the nature of MAL to select a more balanced query, which results in more even performance increases across the classes, and thus a higher macro F1 score for a comparable test accuracy.

One \textbf{\underline{limitation}} of MAL is relatively low performance in the 1-shot setting, which is due to the use of K-Means clustering to circumvent the lack of information in the first cycle. This could be remedied, at the cost of budget, by using an initial seed. In addition, the use of pseudo-labels in MAL is simple and straightforward. A more sophisticated design would potentially improve the performance. For example, one can assign multiple pseudo-labels to each sample, and reduce overlap on the whole set of pseudo-labels, rather than the most likely one. We leave the pursuit of these directions for future work. 

\begin{table*}[t]
\renewcommand{\arraystretch}{1.25}
\centering
\begin{tabular}{cccccccc}
\toprule
\multicolumn{1}{l|}{}& \multicolumn{1}{c}{10-shot (0.11\% Lbl)} & \multicolumn{1}{c}{0.5\% Lbl} & \multicolumn{1}{c}{5\% Lbl} & \multicolumn{1}{c}{20\% Lbl} & \multicolumn{1}{c}{50\% Lbl} & \multicolumn{1}{c}{100\% Lbl} \\ \midrule
\multicolumn{1}{l|}{Acc} & {87.1}\scriptsize$\pm0.5$ & {91.1}\scriptsize$\pm0.5$ & {95.9}\scriptsize$\pm0.3$ & {96.8}\scriptsize$\pm0.1$ & {97.2}\scriptsize$\pm0.1$ & {97.3}\scriptsize$\pm0.0$ \\
\multicolumn{1}{l|}{F1} & {87.3}\scriptsize$\pm0.4$ & {91.2}\scriptsize$\pm0.4$ & {95.4}\scriptsize$\pm0.2$ & {96.9}\scriptsize$\pm0.1$ & {97.3}\scriptsize$\pm0.1$ & {97.4}\scriptsize$\pm0.0$ \\\bottomrule
\end{tabular}
\vspace{5pt}
\caption{MAL in higher annotation budget settings on the NCT dataset. Test accuracies (\%) and macro F1 scores (\%) are shown for all label amounts. For a reference, a fully supervised CNN-based classifier trained on the entire NCT dataset achieves 96.16\% accuracy~\cite{EfficientNet, ghoshnctcnn}.}
\label{table:ssl}
\end{table*}

\begin{table*}[!ht]
\centering
\renewcommand{\arraystretch}{1.25}
\begin{tabular}{cccc|ccc}
\toprule
Pseudo-Labels & SSL Pretraining & Sub-arrays & M-E & 1-shot & 5-shot & 10-shot \\ \midrule
$\times$ & $\times$ & $\times$ & $\times$ & 21.2\scriptsize$\pm5.1$ & 49.5\scriptsize$\pm9.2$ & 59.4\scriptsize$\pm3.5$  \\
$\checkmark$ & $\times$ & $\times$ & $\times$ & 35.3\scriptsize$\pm3.6$ & 55.3\scriptsize$\pm3.5$ & 68.4\scriptsize$\pm2.7$ \\
$\checkmark$ & $\checkmark$ & $\times$ & $\times$ & 42.0\scriptsize$\pm7.5$ & 64.7\scriptsize$\pm2.8$ & 75.6\scriptsize$\pm1.4$ \\
$\checkmark$ & $\checkmark$ & $\checkmark$ & $\times$ & 45.4\scriptsize$\pm6.4$  & 71.2\scriptsize$\pm3.2$ & 78.5\scriptsize$\pm0.2$ \\
$\checkmark$ & $\checkmark$ & $\checkmark$ & $\checkmark$ & \textbf{48.7}\scriptsize$\pm6.7$ & \textbf{77.9}\scriptsize$\pm2.0$ & \textbf{87.1}\scriptsize$\pm0.5$\\ \bottomrule
\end{tabular}
\vspace{5pt}
\caption{Ablation on the components of MAL. The numerical numbers are test accuracy (\%) on the NCT dataset.}
\label{table:ablation}
\end{table*}

\subsection{Ablation Studies}
\label{sec:ablation}

\textbf{MAL Components}: In this ablation study, we gradually remove the four essential components of MAL to measure their impact on the overall performance on the NCT dataset. The four components are pseudo-label generation, the SSL pre-trained encoder, the segmentation of the uncertainty list into sub-arrays, and margin-entropy sampling. These are denoted as pseudo-labels, SSL pretraining, Sub-array, and M-E in Table \ref{table:ablation}, respectively. In each case, the relevant feature is either omitted, or replaced by a conventional version. Specifically, where the SSL encoder is not used, a few-shot learning model pretrained on the CRC-TP dataset (280 000 images, 7-classes of colorectal cancer)~\cite{crctpdataset} is used; and when M-E sampling is not used, the conventional entropy sampling is used instead. As seen in Table \ref{table:ablation}, when each piece of MAL is removed, the test accuracies drop a significant amount at 5 and 10 shots. The 1-shot performance is once again hampered by the lack of target dataset information, but generally decreases as parts are removed. This high variance can be attributed to the lack of knowledge in the early cycles - in the first cycle, the algorithm has no information on the target dataset, so the samples selected are generally low quality, and do not accurately represent the underlying distribution of data.

\section{Conclusions and Future work}
\label{sec:discussion}

In this work, we proposed Myriad Active Learning (MAL), a framework to efficiently and effectively increase the classification accuracy of an active few shot learning model by utilizing unlabelled data. MAL exploits the nature of uncertainty-based active learning sample selection by combining classical uncertainty estimation techniques for a more precise and deliberate query selection strategy. Pseudo-labels generated by an SSL encoder and classifier informed the active learning queries, allowing each sample to be "aware" of the others for a consistently more diverse and less redundant query. MAL produced excellent results, achieving comparable classification accuracy of a fully supervised model on the NCT dataset with only 5\% annotation. MAL also outperforms current few-shot learning methods at 5 and 10 shots, and outperforms common active learning methods in the limited budget setting.

One may notice that in our few-shot active learning paradigm, a SSL encoder is trained on the unlabelled data for the initial pseudo-label generation, avoiding cold start in active learning. Recently, we witness the surge of foundation models. We hypothesize the knowledge in these foundation models would be another good source for pseudo-labels. We will validate this hypothesis in future studies.


{\small
\bibliographystyle{ieee_fullname}
\bibliography{main_cai.bib}
}

\end{document}